\documentclass[conference]{IEEEtran}
\usepackage{amsthm,amssymb,graphicx,multirow,amsmath,color,amsfonts}

\usepackage{subcaption}
\usepackage{caption}
\usepackage[update,prepend]{epstopdf}
\usepackage[noadjust]{cite}

\usepackage{tikz}
\usepackage{bbm} 
\usepackage{multirow}
\usepackage{comment}

\usepackage{algorithm}
\usepackage{algpseudocode}
\usepackage{url}
\def\BibTeX{{\rm B\kern-.05em{\sc i\kern-.025em b}\kern-.08em
    T\kern-.1667em\lower.7ex\hbox{E}\kern-.125emX}}

\allowdisplaybreaks 

\begin{document}
\title{
Deep Learning for Rain Fade Prediction in Satellite Communications}
\author{
Aidin Ferdowsi and David Whitefield\\
Hughes Network Systems, LLC., Emails: \{Aidin.FerdowsiKhosrowshahi,David.Whitefield\}@Hughes.com
}

\maketitle

\begin{abstract}
Line of sight satellite systems, unmanned aerial vehicles, high-altitude platforms, and microwave links that operate on frequency bands such as Ka-band or higher are extremely susceptible to rain. Thus, rain fade forecasting for these systems is critical because it allows the system to switch between ground gateways proactively before a rain fade event to maintain seamless service. Although empirical, statistical, and fade slope models can predict rain fade to some extent, they typically require statistical measurements of rain characteristics in a given area and cannot be generalized to a large scale system. Furthermore, such models typically predict near-future rain fade events but are incapable of forecasting far into the future, making proactive resource management more difficult. In this paper, a deep learning (DL)-based architecture is proposed that forecasts future rain fade using satellite and radar imagery data as well as link power measurements. Furthermore, the data preprocessing and architectural design have been thoroughly explained and multiple experiments have been conducted. Experiments show that the proposed DL architecture outperforms current state-of-the-art machine learning-based algorithms in rain fade forecasting in the near and long term. Moreover, the results indicate that radar data with weather condition information is more effective for short-term prediction, while satellite data with cloud movement information is more effective for long-term predictions.
\end{abstract}
\section{Introduction} \label{sec:intro}
Rain fade refers to the radio signal fade issues caused by rain. The effects of rain fade are more widely seen in satellite systems, unmanned aerial vehicles (UAVs), high-altitude platforms (HAPs), and microwave links that use higher frequency bands, such as Ka-band, Q-band, or V-band. A typical communication process in these systems is composed of four different links: 1) radio frequency (RF) GW (GW) to aerial station (satellite, UAV, or HAP) link, 2) aerial station to remote link, 3) Remote to aerial station link, and 4) aerial station to RF GW link and for each of these links a different mechanism can be implemented to overcome the rain fade. For GW to aerial station link, the trasnmitter in the aerial station is often implemented with automatic power level control that mitigates rain fades to some level.  Moreover, in case of heavy rain fade, automatic uplink power control can be activated to maintain the predefined received power at the the aerial station\cite{Pan2008}. To mitigate the rain fade effect on aerial to remote and remote to aerial links adaptive coding and modulation and adaptive inroute selection algorithms can be used. The aerial to GW link is also susceptible to the effects of rain fade but this is mitigated using the large size and gain of the GW antenna. For ground GW locations which experience extremely high rain fade, an RF GW can connect to a second geographically-separated antenna providing RF terminal (RFT) diversity. The system can automatically switch between the antennas based on the link conditions. However, if the system can predict the occurrence of rain fades, then it can proactively switch between the primary and diversity GWs to maintain the quality of service. Hence, rain fade forecasting is a crucial step in RFT GW diversity switchover and switch back design\cite{samad2021survey}.

The works in \cite{Emilio2008,Lu2018,Mello2009,Xiang2014} have proposed emprical models for predicting the rain fade attenuation based on rain fall rate. In \cite{Emilio2008}, a synthentic storm technique is used that converts a rain rate time series recorded at a give location into a signal attenuation time series. The authors in \cite{Lu2018} have proposed a conversion of rain fall to rain attenuation method which uses the genetic and anealling algorithms to find the parameters for conversion. The work in \cite{Mello2009} has proposed to use an ensemble of several non-linear regression models from different rain fall datasets. Moreover, a path length adjustment is proposed in \cite{Xiang2014} that maps rain fall to rain fade attenuation by taking into account the path length and the frequency of operation. In addition, several statistical models have been proposed in \cite{Gong2013,series2015propagation,Dahman2018} which are based on long-term data of rain attenuation, rainfall rate, and statistical analysis of related atmospheric parameters.

Furthermore, the works in \cite{ITU-R2005,Das2014,Dao2013} have invesitgated the so-called fade slope models. The fade slope model calculates a change in rain attenuation based on fluctuations in measured experimental rain attenuation over time. Following that, the results can be utilized to forecast rain attenuation. Recently, machine learning (ML)-based rain attenuation prediction is proposed in several works \cite{Roy2012,Livieratos2019,Ahuna2019}. Long-term rain attenuation and massive datasets of related parameters such as path length, frequency, rain fall rate, rain drop temperature, humidity, pressure, wind velocity, and etc. are used to train an ML model (i.e., artificial neural network), and this trained model (e.g., optimized weights) can later be used to predict rain attenuation.

However, the atmospheric processes that cause rainfall and subsequently rain fade are complex, and emperical or statistical models in \cite{Emilio2008,Lu2018,Mello2009,Xiang2014,ITU-R2005,Das2014,Dao2013} cannot accurately predict them. These models require field knowledge and sheer data about rain parameters, which can vary dramatically between locations on Earth, thus, making the models difficult to generalize. Although previous works in \cite{Roy2012,Livieratos2019,Ahuna2019} have attempted to address these challenges by using ML models that require the least amount of field knowledge about rain models, they still rely on temporal data on rain parameters and cannot generalize to large scale systems. With the advancement of deep learning (DL) in image processing, recently, many works have proposed deep neural network architectures for precipitation forecasting using satellite and radar images of Earth \cite{Nielsen2021,sonderby2020metnet,prudden2020review}.  Therefore, we use both spatial (radar and satellite images) and temporal (power beacon measurements) to predict chances of rain fade. 

The main contribution of this paper is to propose a novel DL architecture that uses spatio-temporal data from several RF GWs in the Echostar-19 system to classify the satellite-to-GW link status into fade or non-fade classes \cite{echostar-19}. In particular, we have proposed  a 3-D convolutional neural network (CNN) that receives the Geostationary Operational Environmental Satellite 16 (GOES-16) satellite images, Meteorologix radar data, Echostar-19 system link power data as input and extract necessary features to forecast rainfade. We have explained all the preprocessing steps to prepare the data to train the model. To the best of our knowledge, this is the first work that uses weather imagery data to forecast the satellite-to-earth rainfade. Our expereminetal results have shown that the proposed DL architecture can outperform current state of the art works in terms of accuracy, precision, recall, and f1-score in a satellite communication scenario. The proposed DL arhictecture can also be used as a building block of rain fade forecasting for any aerial communication system such as UAVs or HAPs that uses Ka-band or higher frequencies.

\section{Problem Definition}\label{sec:Problem}
GOES-16 provides continuous weather imagery and monitoring of meteorological and space environment data across North America. The satellite provides advanced imaging with high spatial resolution, 16 spectral channels, and with 5 minutes scan frequency for accurate forecasts and timely warnings. The real-time feed and full historical archive of Advanced Baseline Imager (ABI) radiance data (Level 1b) are freely available on Amazon Web Services (AWS) for anyone to use. In addition, every five minutes the Meterologix generates and distributes a file containing a 1 km $\times$ 1 km resolution mosaic of National Weather Service (NWS) radar reflectivity activity. The GOES-16 and Meterologix images cover contiguous United States and the upcoming Echostar-24 GWs are all located within the Western US. Our goal is to design a DL architecture that processes GOES-16 satellite images and Meterologix radar images and the information about rain fade at GWs such as received power from the satellite on its beacon signal received at GWs and forecasts rain fade events in the future.  {It should be noted that the target forecasting time can be as short as the period of beacon records, which is in the order of milli seconds, but in our case, we aim for a 5 minute window due to the model's application in GW switchovers. }

\subsection{Data Preprocessing}
\subsubsection{Downsampling for resolution}
The GOES-16 bucket on AWS contains images of 16 spectral channels (0.47 $\mu m$ - 13.3 $\mu m$) with a 5-minute sampling rate. However, there are some problems with this raw data that need to be addressed. First, the channels have different spatial resolutions varying from 0.000014 to 0.000056 radians in Geostationary coordinate system reference (CSR). Therefore, the channels with a higher resolution need to be downsampled to match with the minimum resolution of the channels or the lower resolution channels need to be upsampled to match with the maximum resolution. The latter case will be less preferable for our problem since upsampling will result in increasing the sizes of the files (and consequently the processing requirements). Thus, the first step in the preprocessing of the AWS files is to downsample the channels with higher resolution from 0.000014 radians to a 0.000056-radian resolution. With 0.000056-radian resolution, every pixel of the image will cover approximately a 2 Km $ \times $ 2 Km area on the US map.
\subsubsection{CRS}
GOES-16 uses Geostationary CRS. Therefore, we will need to transform the GOES-16 coordinates from Geostationary to Geodetic which is the commonly used CRS that describes the location of each GW in latitude and longitude. This conversion is not required for the Meterologix radar data as it is already in Geodetic CRS.
\subsubsection{Extracting the area of interest}
Since we are interested in forecasting the rain fade at the location of GWs, we need to extract the images of areas of interest (AoI) from the raw GOES-16 and radar images. Thus, we crop square areas from the original raw images such that the GWs are located in the center of each square. The number of pixels in these cropped images depends on the size of the area within which we will want to forecast the rain fade event for. For example, a 32 pixels $\times$ 32 pixels image will cover an area of approximately 64 Km $\times$ 64 Km. In fact, our experiments have shown that a 64 Km $\times$ 64 Km AoI results in the highest performance in terms of forecasting the rain fade events. 

\subsubsection{Decomposing Weather Condition Channels}
The values of each pixel in a Meterologix file can take a value between 0 to 48 such that 0 to 16 indicates the intensity of rain, 17 to 32 indicates the intensity of a mix of snow and rain, and 33 to 48 indicates the intensity of snow. Thus we decompes each Meterologix file into three channels of rain, snow, and mix.

\begin{figure*}[t!]
     \centering
     \includegraphics[width=0.75\textwidth]{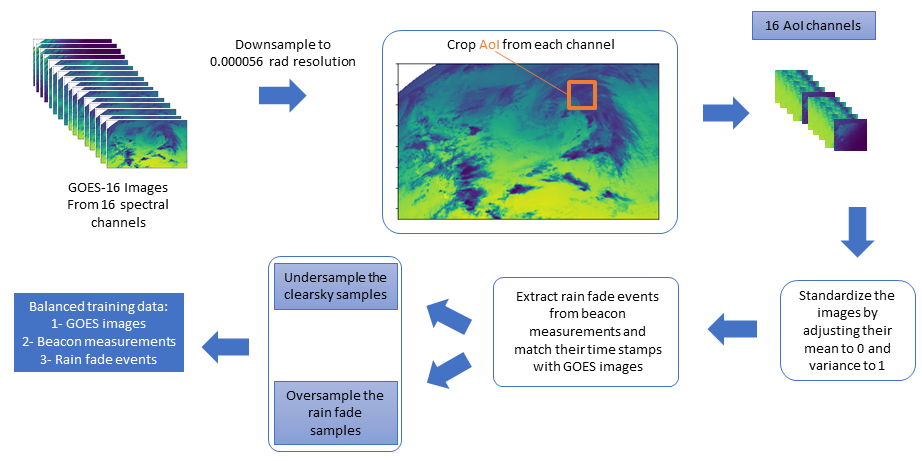}
	\vspace{-3mm}
     \caption{An illustration of data preprocessing steps.}
     \label{fig:preprocessing}
	\vspace{-5mm}
\end{figure*}

\subsubsection{Standardizing the images}
To stabilize our deep learning training, it is recommended to standardize the input data. To this end, the mean value of each channel must be subtracted from the pixels of each channel and then divided by the standard deviation of the channel. This will lead us to have the mean and the standard deviation of the input channels equal to zero and one respectively. Formally, if we let be $p_{ij}^c$ to be the pixel of an image from channel c located at the $i$-th row and $j$-th column, then the standardized pixel will be $\bar{p}_{ij}^c = \frac{p_{ij}^c - m^c}{s^c}$,
where $m^c$ and $s^c$ are the sample mean and the sample standard deviation values of channel $c$. Note that to derive $m^c$ and $s^c$ from the images, we will need to have access to all the images. However, due to the size and number of the image samples, we will use a running approach, particularly Welford's online algorithm \cite{welford1962note}, to calculate and update the mean and standard deviation values. 
\subsubsection{Ground truth label extraction}
For ground truth, we use the beacon measurements and compare them with a rain fade threshold determined by link budget analysis. For training purpose, we sample the beacon data with a 1-minute sampling. For each time sample we derive three values, the minimum beacon value within the past five minutes, the label for the past five minutes, and the label for the future five minutes. For the past or future labels, we consider it to be 1 if the minimum beacon value of past or future 5 minutes goes below the rain fade threshold. For instance, if three consecutive sampling time instances are $t_1$, $t_2$, and $t_3$, then the minimum beacon value between $t_1$ and $t_2$ is used to define the past label at time instance $t_2$ and the minimum beacon value between $t_2$ and $t_3$ is used to define the future label at this time instance. We call the resulting sample and ground truth from the past 5 minute the "current beacon value'' and "current rainfade status'' and the resulting label for the future 5 minutes the "target label''. The current beacon value and current rainfade status will be used in the model as extra information besides the GOES-16 and radar data to improve the accuracy of the model. Note that, although we collect instantaneous beacon samples, the sampling rate for GOES-16 and radar data is 5 minutes, thus, we will not have any image data for 5 minutes after receiving a new GOES-16 or radar data. Hence, we can use the most recent GOES-16 or radar image in case we are to call the model in between two sampling time steps. Therefore, in order to make the model semi-on demand, meaning that instead of waiting for 5 minutes to call the model to predict for the future 5 minutes, in the training phase, we use 1-minute sampling rate which will provide us with more granularity. 

\subsubsection{Imbalanaced data}
The rain fade labels are extremely imbalanced meaning that less than 1\% of the samples can be labeled as rain fade due to the weather condition at these locations. Therefore, using all of the samples will introduce a bias to the model and will increase the number of false negatives (FN) predictions. To address this problem, we will under-sample the clear weather (no rain fade) samples and oversample the rain fade samples to balance the number of samples for true (rain fade) and false (clear) cases. To under sample, we will periodically drop some of the clear sky instances, while for oversampling we use multiple copies of the rain fade instances for the training of the DL architecture. Therefore, by oversampling the rain fade images, and under-sampling the clear sky images we will balance the ratio between the number of true and false samples.

Fig. \ref{fig:preprocessing} summarizes the preprocessing steps of the design. Next, we will propose a DL architecture that can forecast rain fade events based on the preprocessed data.

\section{Deep learning architecture}
\begin{figure*}[t!]
     \centering
     \includegraphics[width=0.8\textwidth]{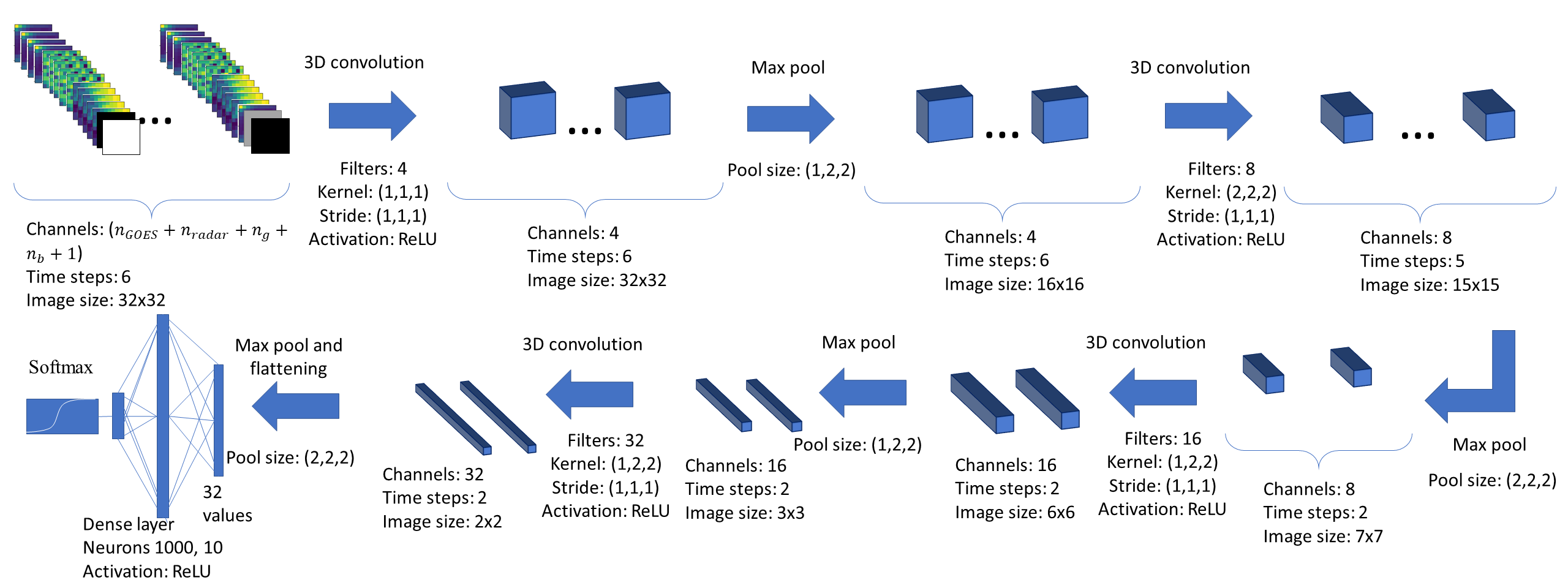}
\vspace{-3mm}
     \caption{The proposed deep learning architecture.}
     \label{fig:architecture}
	\vspace{-6mm}
\end{figure*}

To capture the Spatio-temporal interdependencies of the GOES-16 and radar images we choose a 3D CNN as the main building block for our baseline architecture. A long short term memory (LSTM)-2D CNN may also be evaluated as an alternative in the future.  While a 2D CNN can extract spatial features from an input image, a 3D CNN (or an LSTM-CNN) block can learn the temporal relationship between the input images from different time steps. As shown in Fig. \ref{fig:architecture}, we consider multiple layers of 3D CNN such that conceptually the first layer extracts the interdependencies between the channels and the second or later layers find the features of the images that can be used for rainy weather forecasting. After every CNN layer, we have implemented a pooling layer to reduce the size of the input. CNN utilizes non-linear rectifier (RELU) activation.  And finally, after a flattening layer, we have a dense layer that learns the relationship between the input images and the probability of the rain fades. The activation function of the last layer is chosen to be a softmax to map the output of the dense layer to a rain fade probability value between 0 and 1. In the following, we explain how we prepare the input data for the DL model.

\subsubsection{Preparing the input data for the DL model}
Although GEOS-16 and radar images are the main sources of input for training the DL model, we will attach some extra information to them regarding the location of GWs, the current rainfade state of the GW for each input sample interval, and the current beacon measurement. To this end, based on our experiments, the most efficient way to integrate such information is to add extra channels to the image data in the following way. For the location, we use one-hot encoding meaning that for $n_g$ number of GWs, we consider $n_g$ extra channels. All the pixels of these channels will have a zero value except the i-th channel which will be all ones if the GOES-16 image is for the i-th GW. This input allows multiple GWs to share the same prediction model allowing them generalize better.  Moreover, we add one extra channel such that its pixels will be all +1s if the current state of the GW is rain fade and it will be all -1s, otherwise. This input allows the model to see the ground truth about the rainfade of the GW in the recent past for the given GEOs image at the given time.  Finally, based on the historical beacon data for each GW, we bucketize the beacon measurements into $n_b$ buckets such that each bucket will have approximately equal number of samples. Thus, for each bucket we will have two values that define the two ends of the bucket. Then we consider $n_b$ extra channel such that if the current beacon value falls into the i-th bucket we will define the i-th channel to be all 1s and the other channels to be all -1 (one hot encoding). Thus, considering $n_{\textrm{GOES}}$ and $n_{\textrm{radar}}$ to be the number of channels from GOES-16 and radar sources, at every time step will have $n_{\textrm{GOES}}+n_{\textrm{radar}}+n_g+n_b+1$ channels. Note that if only GOES-16 or radar data is fed to the model, the number channels are chosen to be $n_{\textrm{GOES}}+n_g+n_b+1$ or $n_{\textrm{radar}}+n_g+n_b+1$. Next, the input data for each time step will have a $32 \times 32 \times (n_{\textrm{GOES}}+n_{\textrm{radar}}+n_g+n_b+1 )$ size where 32 is the number of pixels in each direction of the GOES-16 and radar images. In addition, as mentioned, we will also feed the images of the multiple steps in the past to the 3D CNN to capture the temporal behavior of the input images. Thus, letting $n_p$ be the number of samples from past, then input sample to the CNN will have a $n_p \times 32 \times 32 \times (n_{\textrm{GOES}}+n_{\textrm{radar}}+n_g+n_b+1 )$ shape. 

To train the model we split the data into two sets: training and test sets. For the training set, we use the first 80\% of the preprocessed data and keep the remaining 20\% for testing the model. Note that the under-sampling and over sampling steps of the preprocessing are done only on the training set. {Furthermore, we avoid using a validation set; instead, the test set is used to validate the model's performance at each epoch of training.}

\begin{figure}[t!]
     \centering
     \includegraphics[width=0.75\columnwidth]{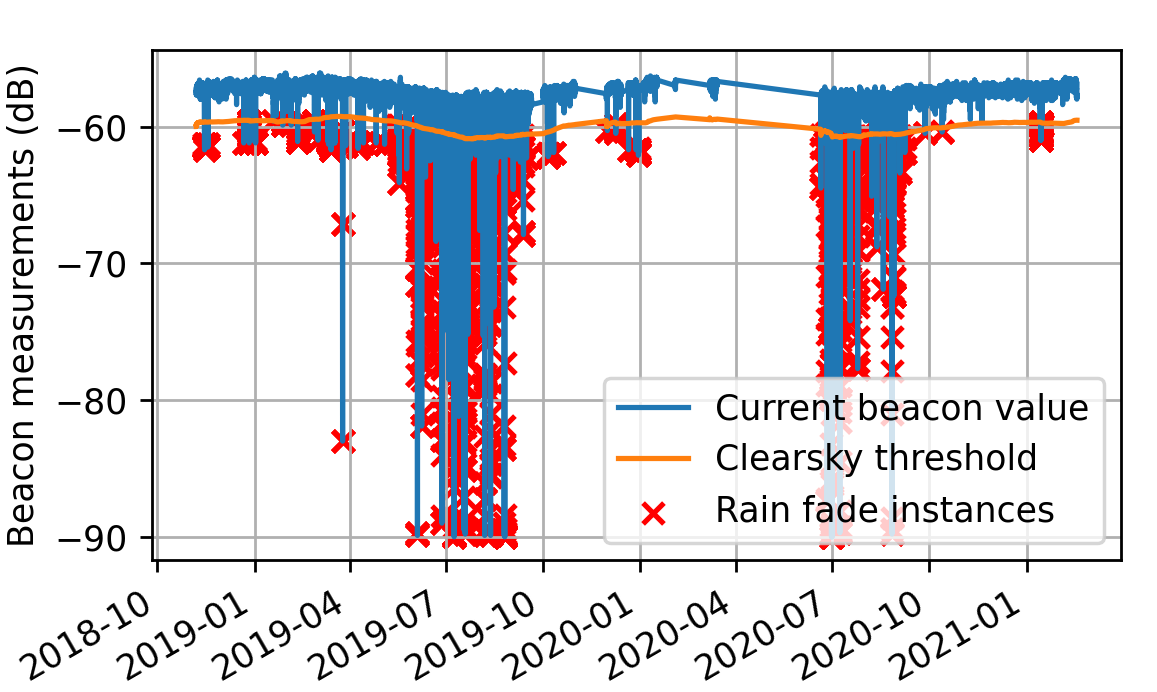}
     \caption{The beacon measurements for a sample GW.}
     \label{fig:beacon}
	\vspace{-5mm}
\end{figure}

\begin{figure*}
     \centering
     \includegraphics[width=0.75\textwidth]{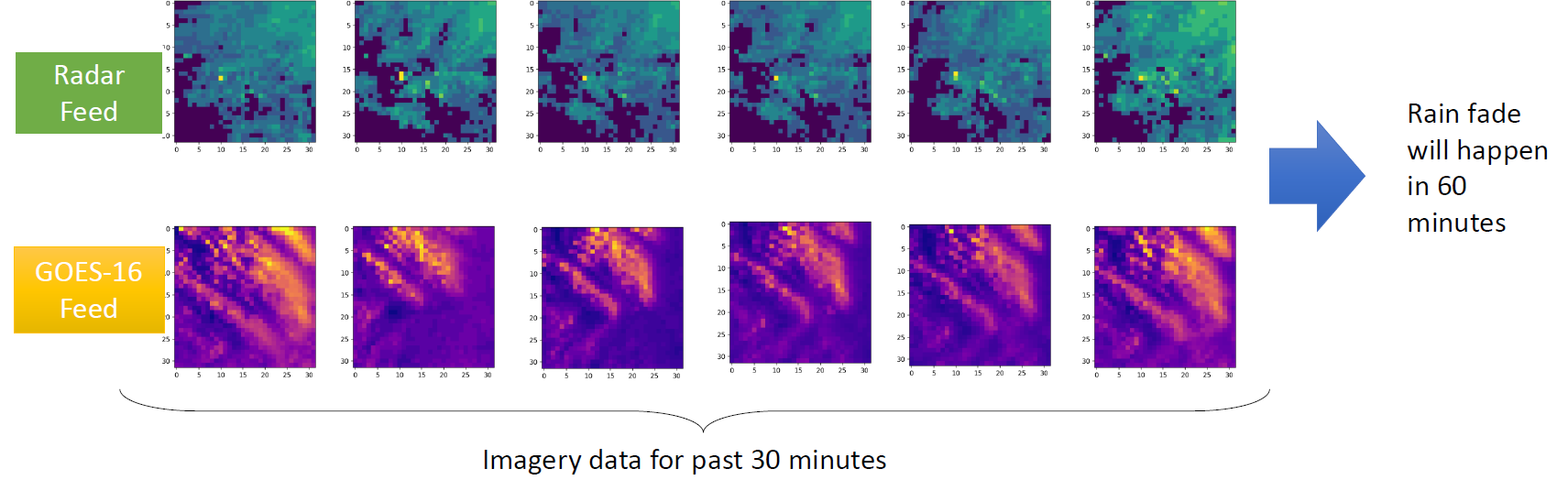}
     \caption{An example of true rain fade prediction for 60 minutes into the future using the imagery data from past 30 minutes.}
     \label{fig:example}
	\vspace{-5mm}
\end{figure*}

\section{Experimental results}
\subsection{Evaluation metrics}
To evaluate the performance of the model let us define four terminologies:
True-positive (TP): A rain fade event correctly classified as rain fade.
False-positive (FP): A clear sky event incorrectly classified as rain fade.
True-negative (TN): A clear sky event correctly classified as clear sky.
False-negative (FN): A rain fade event incorrectly classified as clear sky.
Now, we can define the evaluation metrics:

Accuracy: The closeness of the predictions to their actual labels which is defined as:
\begin{align}
    \textrm{accuracy}=  \frac{(\textrm{TP}+\textrm{TN})}{(\textrm{TP}+\textrm{TN}+\textrm{FN}+\textrm{FP})}
\end{align}

Precision: The fraction of TP instances among the positive instances predicted by the model and defined as:
\begin{align}
    \textrm{precision}=\frac{\textrm{TP}}{(\textrm{TP}+\textrm{FP})}
\end{align}

Recall: The fraction of TP instances among the actual (ground truth) positive instances and defined as:
\begin{align}
    \textrm{recall}=\frac{\textrm{TP}}{(\textrm{TP}+\textrm{FN})}  
\end{align}

F1-score: A harmonic mean of precision and recall which allows us to combine these two metrics and it is defined as:
\begin{align}
    \textrm{F1-score}=2 \frac{(\textrm{precision} \times  \textrm{recall})}{(\textrm{precision}+\textrm{recall})}
\end{align}

F1-score allows us to evaluate the model during the training phase to find a model that has both good precision and recall rates.

\subsection{Dataset}
We choose seven collocated locations for Echostar 19 and Echostar 24 as described in \cite{gateways1} and \cite{gateways2} and use the data from fourth quarter of 2018 till first quarter of 2021.  To evaluate the model first we label the data by aggregating the beacon measurements of each GW and using a weighted avergaing we derive the clear sky threshold for each time step and we compare the beacon measurements of each day by this threshold. Fig. \ref{fig:beacon} shows the beacon measurements, clear sky threshold, and rain fade cases for a single GW.

\begin{figure*}[t!]
     \centering
     \includegraphics[width=0.8\textwidth]{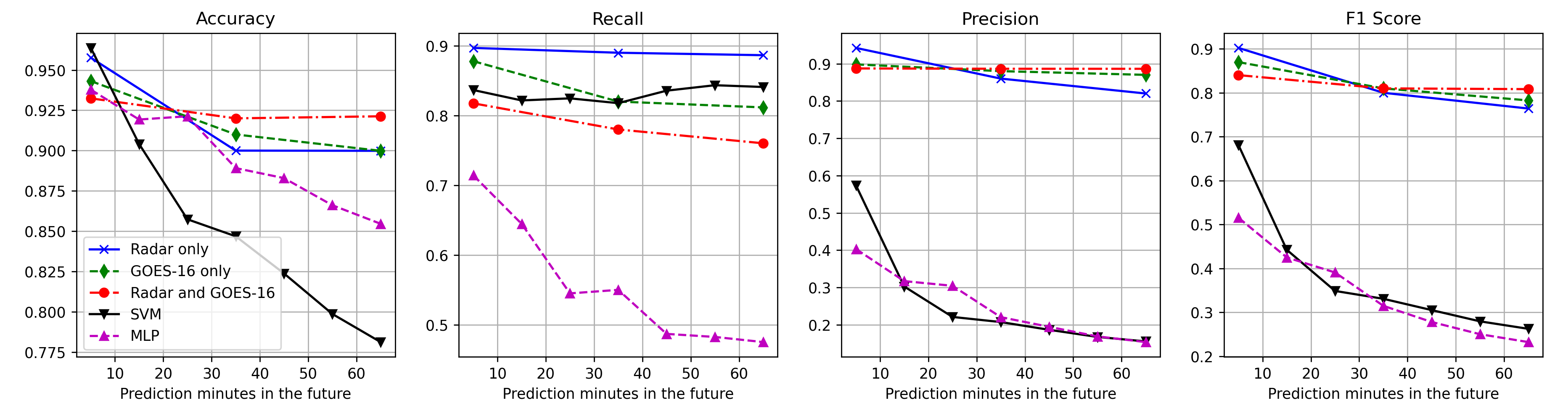}
     \caption{The comparison of our proposed DL architecture with state-of-the-art ML models.}
     \label{fig:comparison}
\vspace{-5mm}
\end{figure*}

\subsection{Experiments}

Fig. \ref{fig:example} shows an example of the input imagery from past 30 minutes to the model and its correct prediction of rain fade in 60 minutes into the future. In Fig. \ref{fig:example}, {we show an example of the model's data inputs and how the model can predict a long-term rain fade event in 60 minutes by processing data from the previous 30 minutes.} This is a result that other state-of-the-art models fail to achieve. To showcase this performance gain, next, we train our model for different target future time from 5 minutes to 65 minutes into the future. 

First, we train our proposed model on three imagery input scenarios: a) GOES-16 only, b) radar only, and c) GOES-16 and radar together. In addition to the imagery data, as previously discussed, we use the beacon data as additional information. Fig. \ref{fig:comparison} shows the output of the model on the test data of one RF GW. We can see from Fig. \ref{fig:comparison} that the case with only radar inputs outperforms the other two scenarios for short term forecasting in terms of f1-score while the model trained only on GOES-16 outperforms the other two in long term forecasting. This is due to the fact that GOES-16 images track the movements of the clouds while the radar images have the weather condition records. Thus, for a short term prediction radar data is more effective while for a long term prediction a data fusion of radar and GOES-16 can be used. Next, we compare our model's performance to that of two other cutting-edge ML-based rain fade prediction models that only use time series data. In \cite{Ahuna2019} a multi-layer perceptron (MLP) was used while in \cite{Livieratos2019} a support vector machine (SVM)-based model was proposed for rain fade prediction. In our experiments, we use the beacon data as the time series input for these models. We can see from Fig. \ref{fig:comparison} that the proposed DL architecture can outperfom the other state-of-the art ML models especially for long term predictions.

\begin{figure}[t!]
     \centering
     \includegraphics[width=0.78\columnwidth]{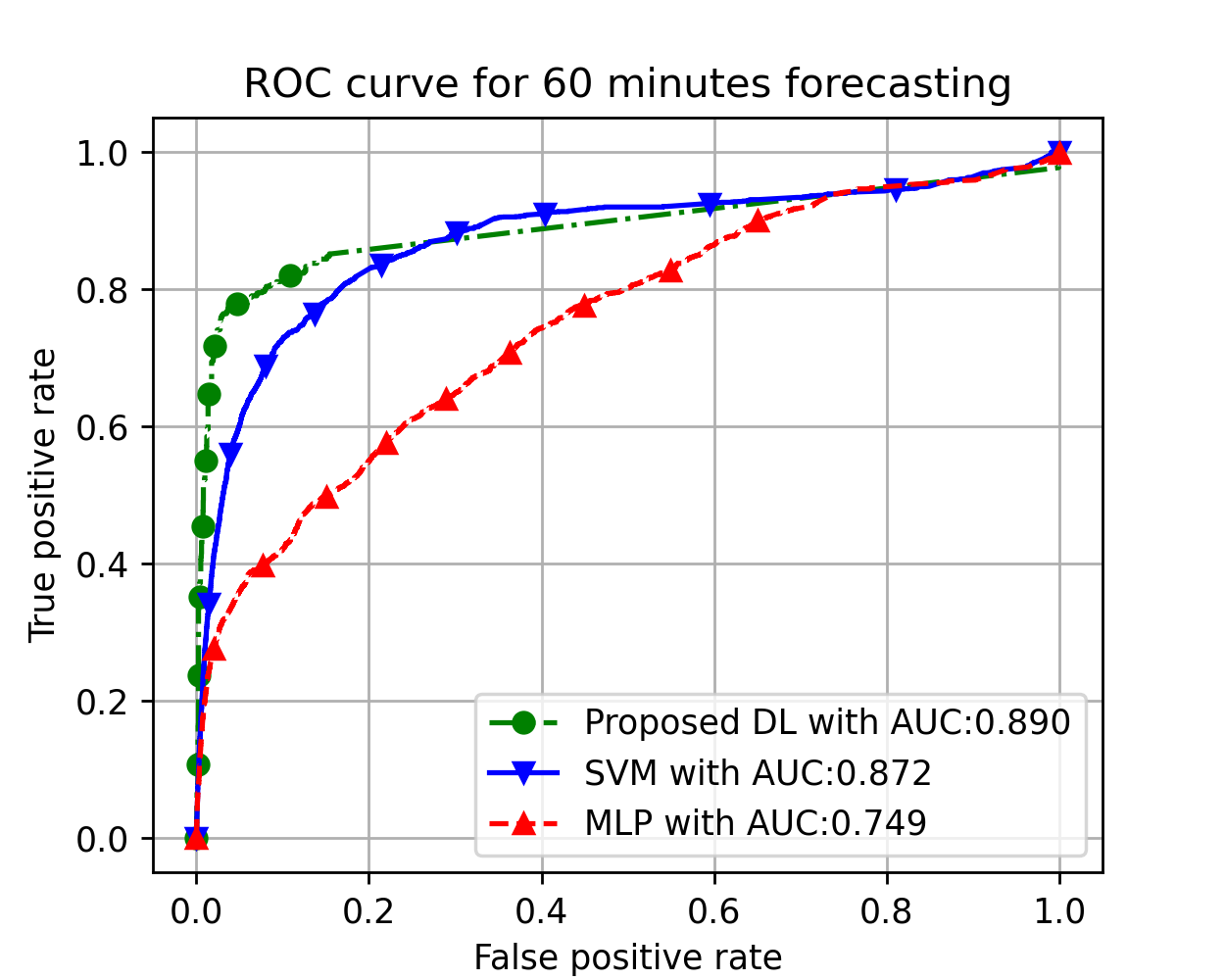}
     \caption{ROC curve of the proposed DL architecture vs state-of-the-art ML models.}
     \label{fig:ROC}
\end{figure}

\begin{figure}[t!]
     \centering
     \includegraphics[width=0.83\columnwidth]{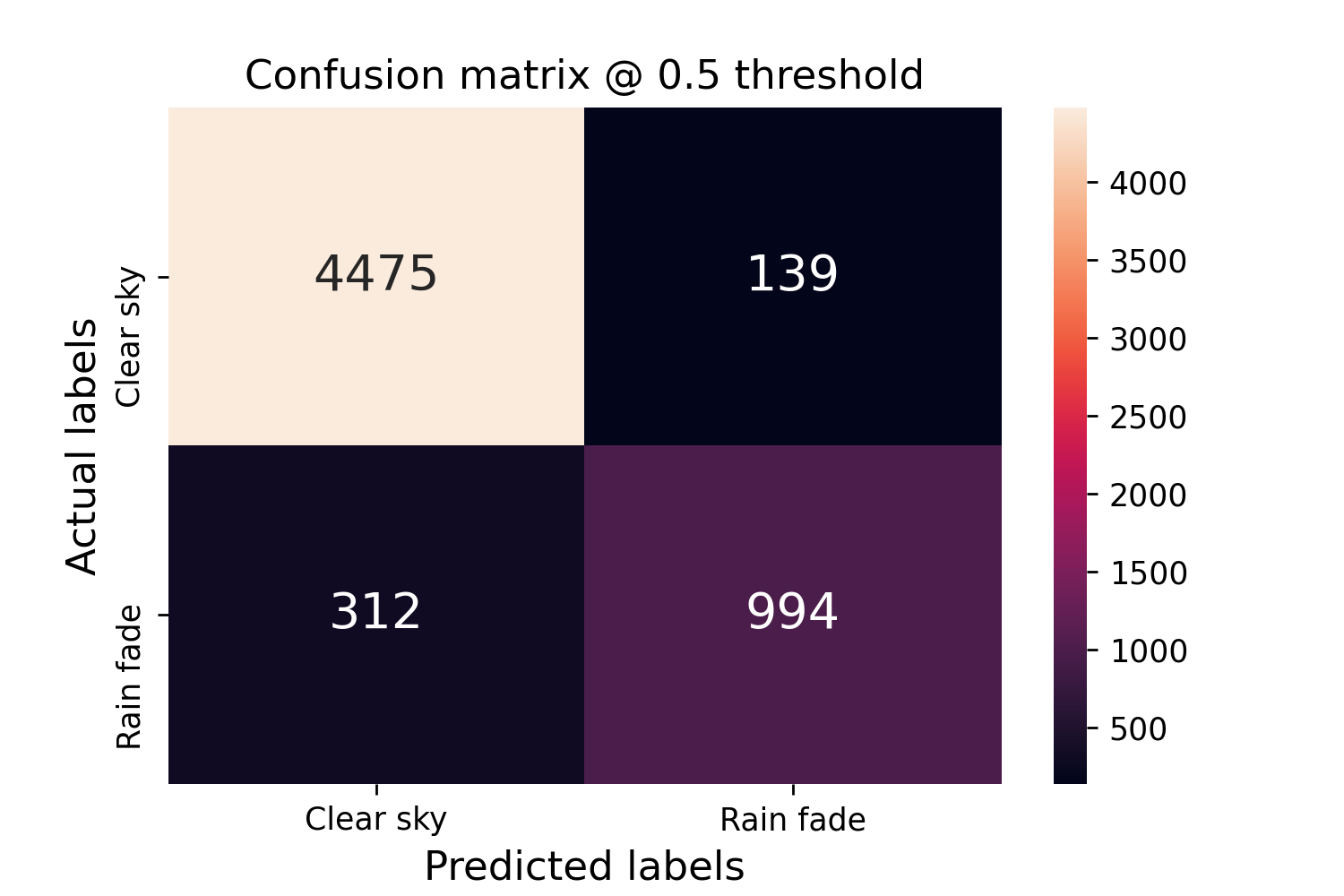}
     \caption{The confusion matrix for the proposed DL when predicting 60 minutes into the future.}
     \label{fig:CM}
\vspace{-3mm}
\end{figure}

Fig. \ref{fig:ROC} shows the receiver operating characteristic (ROC) curve of a long term prediction scenario for our proposed DL model vs the two ML-based models. The ROC curve depicts a trade-off between the TP rate (TPR) and the FP rate (FPR) by plotting TPR versus FPR at various thresholds. Lowering the classification threshold causes more observations to be classified as positive, increasing the TP rate. We can see from Fig. \ref{fig:ROC}, because the ROC curve in our proposed model is closer to the top left of the graph, we can achieve a high TPR while maintaining a low FPR. The other two classifiers, particularly the MLP classifier, will be unable to distinguish between the two classes well, and thus its ROC curve will be closer to the diagonal, implying lower TPR and higher FPR. The area under the ROC Curve (AUC) measures a model's performance across all possible classification thresholds. From Fig. \ref{fig:ROC} we can see that our proposed DL architecture has a higher AUC thus it is better at predicting the probability of rain fade higher than the probability of clear sky. In addition, Fig. \ref{fig:CM} depicts the proposed DL's confusion matrix when predicting rain fade 60 minutes in the future. According to Fig. \ref{fig:CM}, with a classification threshold of 0.5, the proposed architecture accurately predicts rain fade and clear sky events almost 12 times more than the false labels ($\frac{(\textrm{TP}+\textrm{TN})}{(\textrm{FN}+\textrm{FP})} \approx 12$) which shows how effective the architecture is working in terms of forecasting the rain fade.

\section{Conclusion}
In this paper, we have proposed a DL-based architecture for forecasting rain fade in aerial communication systems that operate at Ka-band or higher frequencies (Q-band and V-band), such as satellite systems, UAVs, or HAPs. To that end, we have described all of the steps for preprocessing and preparing the data, as well as the architectural design. We have conducted several experiments for long-term and short-term rain fade forecasting and compared our results to cutting-edge ML-based approaches. The results demonstrate that the proposed DL-based architecture outperforms state-of-the-art models in terms of accuracy, recall, precision, and f1-score, particularly in long-term forecasting. In addition, the experiments show that radar data containing radar weather condition information is more effective for short-term prediction, whereas satellite weather data containing cloud movement information is more effective for long-term prediction.
\bibliographystyle{IEEEtran}
\bibliography{reference}
\end{document}